\documentclass[conference]{IEEEtran}
\IEEEoverridecommandlockouts
\usepackage{cite}
\usepackage{amsmath,amssymb,amsfonts}
\usepackage{algorithmic}
\usepackage{graphicx}
\usepackage{textcomp}
\usepackage{xcolor}
\def\BibTeX{{\rm B\kern-.05em{\sc i\kern-.025em b}\kern-.08em
    T\kern-.1667em\lower.7ex\hbox{E}\kern-.125emX}}

\usepackage{amsmath}
\usepackage{amssymb}
\usepackage{amsthm}
\usepackage{graphicx}
\usepackage{algorithm}
\usepackage[numbers, sort&compress]{natbib}

\newcommand{\n}{\boldsymbol{n}}

\newcommand{\z}{\mathbf{z}}

\newcommand{\E}{\mathbf{E}}

\newcommand{\1}{\mathbf{1}}

\newcommand{\x}{\boldsymbol{x}}

\renewcommand{\log}[1]{{\rm{log}}#1}

\newtheorem{definition}{Definition}

\title{Learning to Detect with Constant False Alarm Rate}

\author{Tzvi Diskin, Uri Okun and Ami Wiesel }
\date{February 2022}
\begin{document}

\maketitle

\begin{abstract}
    We consider the use of machine learning for hypothesis testing with an emphasis on target detection. Classical model-based solutions rely on comparing likelihoods. These are sensitive to imperfect models and are often computationally expensive. In contrast, data-driven machine learning is often more robust and yields  classifiers with fixed computational complexity. Learned detectors usually provide high accuracy with low complexity but do not have a constant false alarm rate (CFAR) as required in many applications. To close this gap, we propose to add a term to the loss function that promotes similar distributions of the detector under any null hypothesis scenario. Experiments show that our approach leads to near CFAR detectors with similar accuracy as their competitors.
\end{abstract}


\section{Introduction}

The deep learning revolution has led many to apply machine learning methods to classical problems in all fields including statistics. Examples range from estimation \cite{dong2015image, ongie2020deep, gabrielli2017introducing, dua2011artificial,dreifuerst2021signalnet,diskin2021learning} to detection \cite{samuel2019learning,girard2021deep,brighente2019machine,de2017approximating,ziemann2018machine,theiler2021bayesian}. Deep learning is a promising approach for deriving high accuracy and low complexity alternatives when classical solutions are intractable. To facilitate this switch, we must ensure that the learned solutions are accurate but also satisfy other classical requirements. In this paper, we focus on learning detectors for composite hypothesis testing. We claim that current solutions deliver on their promises but lack the Constant False Alarm Rate (CFAR) requirement which is critical in many applications. To close this gap, we provide a framework for learning accurate CFAR detectors. 

%Hypothesis testing and target detection are fundamental problems in statistical signal processing. Applications range from radar and imaging to communication systems. 

%Traditionally, hypothesis testing and target detection were solved using tools that choose the hypothesis which is more likely. These model-driven tools led to high accuracy but were often computationally intensive. On the other hand, machine learning considers low-cost classifiers that are fitted to minimize data-driven loss functions. Modern testing problems are somewhere in the middle: it is common to assume a partially-specified and physically-based statistical model, and there are often computational constraints on the final detector. Therefore there is a natural trend to switch to learned detectors. These usually deliver on their promise, but lack the CFAR property which is critical in many applications. 

%In the first part of this paper, we consider the use of machine learning for simple hypothesis testing and review state of the art. The main conclusion is that machine learning provides a promising alternative to classical methods. In the second part, we move on to composite hypothesis testing with unknown parameters. In these more challenging problems we claim that existing methods are insufficient and propose alternative solutions that provide the best of both worlds. 

Detection theory begins with simple hypothesis testing where a detector must decide between two fully specified distributions. The classical solution is the Likelihood Ratio Test (LRT) which is optimal in terms of maximizing the detection probability subject to a false alarm constraint. Composite hypothesis testing is a more challenging setting where the hypotheses involve unknown deterministic parameters. A CFAR detector is invariant to these parameters and has identical false alarm probabilities as long as no target is present. This allows the user to set the thresholds a priori. A popular approach is the Generalized Likelihood Ratio Test (GLRT) which can be interpreted as estimating the unknown parameters and then plugging them into a standard LRT. GLRT performs well when the number of unknowns is relatively small and is asymptotically CFAR. On the other hand, it is generally sub-optimal in finite sample settings and may be computationally expensive. 

Data-driven classifiers are the machine learning counterpart to model-based detectors. In simple settings with no unknown deterministic parameters, it is well known that the optimal Bayes classifier converges to the LRT with a specific false alarm rate. More advanced classifiers can also maximize the cumulative detection rate over a wide range of false alarms, also known as the area under the curve (AUC) \cite{herschtal2004optimising, brefeld2005auc}. 

There is a growing body of works on using machine learning for target detection. In the context of hyperspectral imagery, \cite{ziemann2018machine} introduced the use of 
support vector machines (SVM). Deep neural networks (DNN) were proposed in \cite{girard2021deep}. An important ingredient of these works is the use of an artificial training set where real noise data is augmented with synthetically planted targets. 
  In the context of radar detection, SVMs were considered in \cite{de2017approximating}. Specific CFAR radar detectors were developed by relying on CFAR features \cite{lin2019dl,akhtar2018neural,akhtar2021training}.

The main contribution of this paper is a framework for learning CFAR detectors denoted by CFARnet. The framework is general purpose and can be applied to arbitrary composite hypothesis testing problems. CFARnet is based on adding a penalty function to the optimization which decreases the distances between the distributions under different parameter values. To optimize CFARnet we rely on empirical and differentiable distances that have recently become popular in unsupervised deep learning. Our numerical experiments, show that the resulting networks are approximately CFAR while paying a negligible price in terms of accuracy.

%We perform intensive numerical experiments comparing these approaches in the context of target detection. Our results show that both methods perform similarly and succeed in approximating the LRT uniformly. There are small differences depending on the settings but these are insignificant. 

The CFAR notion is closely related to the topics of ``fairness'' and ``out of distribution (OOD)'' generalization which have recently attracted considerable attention in the  machine learning literature, e.g., \cite{arjovsky2019invariant,wald2021calibration}. 
CFAR can be interpreted as a fairness property with respect to the unknown deterministic parameters. The closest work is \cite{romano2020achieving} which also enforces ``equalized odds'' using a distance between distributions. A main difference is that CFAR is a one-sided fairness property and requires equal rates only in the null hypothesis. %Also, unlike these data-driven methods, CFARnet is based on a well defined generative model with deterministic unknown parameters. 
Algorithmically, \cite{romano2020achieving} compares the high dimensional joint distribution of the predictions and the unknown parameters, whereas we only consider the scalar distribution of the predictions. This makes our method significantly cheaper in terms of computational complexity. Indeed, we rely on a simple kernel based distance and do not require sophisticated adversarial networks.

%Fair learning tries to eliminate biases in the training set and considers properties that need to be protected \cite{agarwal2019fair,bagnell2005robust}. OOD works introduce an additional ``environment'' variable and the goal is to train a model that will generalize well on new unseen environments \cite{creager2021environment,maity2020there,arjovsky2019invariant,wald2021calibration}. 

%Outline ... Notations...

%Another downside to LRT is when the likelihoods are not a priori known and must be estimated from data. In this case, alternative solutions may be advantageous. In this paper, we focus on signal processing applications where an exact physics based model is known. In Section ?, we will also consider partially known models. Fully data-driven hypothesis testing is outside the scope of this paper. 

\section{Background on statistical distances}\label{sec_dist}
We begin with a brief background on statistical distances.

\begin{definition}
Let $X\sim p(X)$ and $Y\sim p(Y)$ be two random variables. A statistical distance $d(X;Y)$ is a function
that satisfies $d(X;Y)\geq 0$ with equality if and only if $p(X)=p(Y)$.
\end{definition}

A distance which has recently become popular is the Maximum Mean Discrepancy (MMD) \cite{gretton2012kernel}:
\begin{multline}\label{mmd_def}
    d_{\rm{MMD}}(X;Y) = \E[k(X,X')] \\ + \E[k(Y,Y')] - 2 \E[k(X,Y)]
\end{multline}
where $X$ and $X'$ are independent and identically distributed (i.i.d.), and so are $Y$ and $Y'$. The function $k(\cdot,\cdot)$ is a characteristic kernel over a reproducing kernel Hilbert space, e.g.,  the Gaussian Radial Basis Function (RBF).

Recent advances in deep generative models allow us to optimize distances as MMD in an empirical and differentiable manner. For this purpose, we need to represent each distribution using a small dataset. Let $\{X_i\}_{i=1}^N$ and $\{Y_i\}_{i=1}^N$ be i.i.d. realizations of $X$ and $Y$, respectively. Then, an empirical version of the MMD can be used where the expetations in (\ref{mmd_def}) are replaced by their empirical estimates.
More advanced metrics can be obtained using  
% \begin{align}
%     \hat R_{\rm{adv}}(\hat{T})= \sum_{i=1}^N\max_{g(\cdot)}\left|\sum_{j=1}^Mg(T_{ij})-\sum_{j=1}^Mg(\tilde{T}_{ij})\right|^2
% \end{align}
the tools of generative adversarial networks (GANs). In this paper, we only deal with distances between scalar random variables and simple MMD distances suffice.  %\cite{goodfellow2014generative,arjovsky2017wasserstein}. 

\section{Problem formulation}
We consider a binary hypothesis test. Let $\x$ be an observed random vector whose distribution $p(\x;\z)$ depends on an unknown deterministic parameter $\z$. The value of $\z$ defines two possible hypotheses
\begin{align}\label{comp_testing}
    &y=0:\quad \z\in{\mathcal{Z}}_0\nonumber\\
    &y=1:\quad \z\in{\mathcal{Z}}_1.
\end{align}
The goal is to design a detector $\hat y(\x)\in\{0,1\}$ as a function of $\x$ that will identify the true hypothesis $y\in\{0,1\}$. 
Performance is measured in terms of probability of correct detection, also known as True Positive Rate (TPR):
\begin{align}\label{PD_FA1}
   P_{\rm TPR}(\z) =  \; &P(\hat y(\x) =1 ;y=1)
\end{align}
and probability of false alarm, also known as False Positive Rate (FPR):
\begin{align}\label{PD_FA2}
   P_{\rm FPR}(\z) =  P(\hat y(\x) =1 ;y=0)
\end{align}
In practice, the user typically provides a false alarm constraint $P_{\rm FPR}\leq \alpha$ that must be satisfied and the goal is to maximize $P_{\rm TPR}$.

It is standard to consider detectors of the form
\begin{equation}\label{detector}
    \hat{y}\left(\x\right)=\begin{cases}
0 & T\left(\x\right)<\gamma\\
1 & T\left(\x\right)\geq\gamma
\end{cases}
\end{equation}
where $T(\x)$ is a function of the measurements and $\gamma$ is a threshold. This structure   allows users to tune $P_{\rm FPR}$ by adjusting the threshold.  Performance is usually visualized using the Receiver Operating Characteristic (ROC) which plots the TPR as a function of the FPR. In signal processing applications, users are often interested in a region of very low FPRs, e.g., $10^{-1}-10^{-3}$ and the goal is to maximize the TPR probabilities in this area.

A main challenge in detection theory are the unknown parameters under the null hypothesis $y=0$. The false alarm probability (FPR) is generally a function of these parameters and cannot be controlled without their knowledge. Therefore, it is often preferable to restrict the attention to CFAR detectors.
\begin{definition}
A detector $T(\x)$ is CFAR if its distribution is invariant to the value of $\z\in{\mathcal{Z}}_0$.
\end{definition}

As we will review below, many classical  detectors are CFAR or  asymptotically CFAR. With the growing trend of switching to machine learning, the goal of this paper is to introduce a framework for learning CFAR detectors.

\section{Classical Likelihood based Detectors}
In this section, we provide a short background on classical detectors based on likelihood ratios. In the simple case where ${\mathcal{Z}}_0=\{\z_0\}$ and ${\mathcal{Z}}_1=\{\z_1\}$ are singletons, hypothesis testing has an optimal solution known as the Likelihood Ratio Test (LRT) due to Neyman Pearson \cite[p. 65]{kay1998fundamentals}. LRT theory states that the optimal detector for maximizing detection subject to a given false alarm probability is:
\begin{align}\label{LRT}
    T_{\rm LRT}(\x) &=2\log\frac{p(\x;\z=\z_1)}{p(\x;\z=\z_0)}
\end{align}
and the threshold $\gamma$ is chosen to satisfy the false alarm (FPR) constraint.

The more realistic scenario is composite hypotheses testing where one or both of the hypotheses involve unknown parameters and there is no simple and optimal solution. A popular heuristic is the Generalized Likelihood Ratio Test (GLRT) that estimates the unknown parameters using the Maximum Likelihood technique and plugs them into the  LRT detector:
\begin{align}\label{GLRT}
    T_{\rm GLRT}(\x) &=  2\log\frac{\max_{\z\in{\mathcal{Z}}_1}p(\x;\z)}{\max_{\z\in{\mathcal{Z}}_0}p(\x;\z)}
\end{align}
Setting the threshold to ensure a fixed $P_{\rm{FPR}}$ is non trivial. Fortunately, under regular conditions, GLRT is asymptotically CFAR and thus its threshold can be set for all values of the unknown parameters simultaneously.
% \begin{align}\label{fim}
%     &\lambda=(\z_{r1}-\z_{r0})^T\F(\z_{r1}-\z_{r0})\nonumber\\
%     &\F = \I_{\z_r\r_r}-\I_{\z_r\r_s}\I^{-1}_{\z_s\r_s}\I_{\z_s\r_r}
% \end{align}
% where $\z_{r1}$ is the ground truth parameter and the $\I$ matrices denote the different sub-blocks of the  Fisher Information Matrix (FIM) as detailed in \cite[6.5]{kay1998fundamentals}. Thus, GLRT is asymptotically CFAR and its threshold can be set using (\ref{asympP}).

%Thisleads to a closed form threshold which guarantees an asymptotic constant false alarm rate (CFAR) for any nuisance parameter $\z_s$.

GLRT is probably the most popular solution to composite hypothesis testing. It gives a simple recipe that performs well under asymptotic conditions. Its main downsides are that it is sensitive to deviations from its theoretical model, it is generally sub-optimal under finite sample settings and that it may be computationally expensive. Both the nominator and denominator of the GLRT involve optimization problems that may be large scale, non-linear and non-convex. Therefore, there is an ongoing search for robust and low cost alternatives. 

% In the classic statistics framework, the goal is to find a test $T(\x)$ and a threshold $\gamma$ such that the detector $\hat \z(\x)$ is given by:

% The performance of the the detector is measured by probability of the detection (TPR), with a constrain on the probability of the false alarm (FPR), that is:
% \begin{align}\label{PD_FA}
%   {\rm TPR} =  \; &p(\hat y(\x) = 1 |y=1) \nonumber \\
%       {\rm s.t.} \;  &p(\hat y(\x) = 1 |y=0) \leq t,
% \end{align}
% for every $0\leq t\leq 1$. The probability of detection as a function of the (FPR) is called the Receiver Operating Characteristic (ROC) curve.

% \subsection{Bayesian Solution}
% \anote{im wondering if we want this section, as its a solution to a different problem}

% An alternative solution to hypothesis testing is the Bayesian framework which assumes random labels $y$ with a known prior $p(y)$. Performance is measured by the probability of error, also called the Bayes risk or the 0-1 loss :
% \begin{equation}\label{prob error}
%     p_{\rm err}=p(\hat y(\x)=1|y=0)p(y=0) + p(\hat y(\x)=0|y=1)p(y=1)
% \end{equation}
% The optimal Bayes detector for minimizing  $p_{\rm err}$ is identical to the LRT (\ref{LRT}), with a threshold that depends on the priors:
% \begin{equation}
%     \gamma = \frac{p(y=0)}{p(y=1)}.
% \end{equation}

\section{Machine Learning for Detection}\label{ml4lrt}

%The recent deep learning revolution has led many to apply machine learning methods to classical problems in statistics, ranging from estimation \cite{dong2015image, ongie2020deep, gabrielli2017introducing, rudi2020parameter, dua2011artificial,dreifuerst2021signalnet} to detection \cite{samuel2019learning,girard2021deep,brighente2019machine,de2017approximating,ziemann2018machine}. Deep learning relies on a computationally intensive fitting phase which is done offline, and yields a neural network with fixed complexity that can be easily applied in inference time. Therefore, it is a promising approach for deriving low complexity detectors when GLRT is intractable. 

In this section, we explain the use of machine learning for hypothesis testing. The starting point to any data-driven learning is a training set. Hypothesis testing relies on a probabilistic model $p(\x;\z)$ and we need to use this model in order to generate a synthetic dataset. Hybrid settings involving a mixture of real and artificial samples are also common. For example, it is standard to plant synthetic targets on real noise samples  \cite{ziemann2018machine}.
Due to space limitations, we leave these hybrid extensions for the journal version of this paper.

The main challenge in generating data is that $y\in\{0,1\}$ and $\z$ are not random variables but deterministic parameters without any prior distribution. A natural heuristic is to assume uniform fake priors, e.g., choose half of labels as $y=0$ and half as $y=1$ and assume that $\z$ is uniformly distributed on its corresponding domains. For each $y_i$ and $\z_i$, we then generate a measurement $\x_i$ according to the true $p\left(\boldsymbol{x};{\z_i}\right)$ and obtain a synthetic dataset
\begin{equation}
    {\mathcal{D}}_N=\{\boldsymbol{x}_i,\z_i,y_i\}_{i=1}^N.
\end{equation}

Next, a class of possible detectors ${\mathcal{T}}$ is chosen in order to tradeoff expressive power with computational complexity in test time. The class is usually a fixed differentiable neural network architecture. In our context, it also makes sense to reuse existing ingredients from classical detector as non-linear features or internal sub-blocks.

Finally, the learned detector is defined as the minimizer of an empirical loss function
\begin{equation}\label{classifcation}
   \min_{\hat{{T}}\in {\mathcal{T}}} \frac 1N \sum_{i=1}^N L(\hat T(\x_i),y_i).
\end{equation}
where $L(\cdot;\cdot)$ is a classification loss function. Ideally, we would like to minimize the zero-one loss which corresponds to the average probability of error. Practically, for efficient optimization, a smooth and convex surrogate loss, as the hinge or cross entropy functions, is preferable. The overall procedure for learning a detector is summarized in Algorithm 1.

\begin{algorithm}
\caption{Fitting detectors for $p(\x;\z)$}\label{alg:BCE}
\begin{itemize}
    \item Choose   $p^{\rm{fake}}(y)$.
    \item Choose  $p^{\rm{fake}}(\z;y)$.
    \item For each $i=1,\cdots,N$:
    
    \hspace{4mm}Generate $y_i$.
    
    \hspace{4mm}Generate $\z_i$ given $y_i$.
    
    \hspace{4mm}Generate $\x_i$ given $\z_i$.
    \item Solve 
    
    \hspace{4mm}$\min_{\hat{{T}}\in {\mathcal{T}}} \frac 1N \sum_{i=1}^N L(\hat T(\x_i),y_i)$.
\end{itemize}
\end{algorithm}

Learned detectors have strong theoretical guarantees in simple testing problems where $y$ uniquely defines $\z$. With a sufficiently expressive class of detectors and a large enough training set, minimizing the zero-one loss leads to the Bayes optimal detector. It is identical to the LRT with the threshold
\begin{equation}
    \gamma = \frac{p^{\rm fake}(y=0)}{p^{\rm fake}(y=1)}.
\end{equation}
Minimizers of Bayes risk consistent surrogates, as the hinge loss,  also asymptotically converge to this LRT \cite{bartlett2006convexity}. Thus, any simple LRT can be approximated by tuning $p^{\rm fake}(y)$ to achieve a desired $P_{\rm FPR}$. 

To approximate LRT for a wide range of $P_{\rm FPR}$ it is preferable to minimize the AUC. There are many works in the machine learning literature on fitting classifiers with this goal through an increased computational complexity \cite{brefeld2005auc, herschtal2004optimising}. There are also variants that focus on partial regimes in the ROC \cite{narasimhan2013structural}. 

%In composite hypothesis testing, learned detectors are heuristics without solid theory. Their performance depends on the choice of $p^{\rm fake}(\z;y)$. To our knowledge, there is no clear relation between Algorithm 1 and GLRT.

% The obvious choice for the loss function is the zero-one loss
% \begin{align}\label{prob error}
%     L_{0-1}(\hat T(\x),\z)=&\left\{\begin{array}{ll}
%          & \hat T(\x)> \frac 12 , H=0 \\
%          1 & \qquad{\rm or}\\
%          & \hat T(\x)\leq  \frac 12 , H=1\\
%         0 & {\rm else}
%     \end{array}\right.
%     % p\left(\hat T(\x)> \frac 12;H=0\right)p^{\rm{fake}}(H=0) \nonumber\\
%     % &+ p\left(\hat T(\x)\leq \frac 12;H=1\right)p^{\rm{fake}}(H=1)
% \end{align}

To evaluate the machine learning approach to detection we turned to numerical experiments. We completed a wide range of simulations in different target detection scenarios comparing the classical solutions with different learned detectors based on various loss functions. We examined both simple and composite settings, with and without hidden variables. The conclusions were that learned detectors usually perform similarly to the their corresponding (G)LRTs. In some experiments, some detectors had a small advantage but the differences were not uniform nor significant. Our conclusion is therefore that simple classifiers with wide priors on the unknown parameters are practically sufficient for achieving an optimal ROC in most problems. On the negative side, most of the experiments showed that the learned detectors were not CFAR and resulted in significantly different false alarm rates for different values of $\z\in{\mathcal{Z}}_0$. To close this gap, in the next section we propose a framework for learning CFAR detectors.

\section{Learning CFAR detectors}

% \begin{figure*}[h]
% \centering
% %\captionsetup{justification=centering}
% \begin{array}{cc}
%      \includegraphics[width=0.3\textwidth]{plots/sigma05_Gaussain.png}\includegraphics[width=0.3\textwidth]{plots/sigma1_Gaussain.png}\includegraphics[width=0.3\textwidth]{plots/fpr_fprGaussain.png} \\
%      \includegraphics[width=0.3\textwidth]{plots/sigma05_non_Gaussain.png}\includegraphics[width=0.3\textwidth]{plots/sigma1_non_Gaussain.png}\includegraphics[width=0.3\textwidth]{plots/fpr_fprnon_Gaussain.png}
% \end{array}
% \caption{Performance graphs in terms of FPR, TPR and thresholds for $\sigma\in\{0,1\}$. Top row is for Gaussian noise, whereas bottom row is for non-Gaussian noise. In terms of ROC, the learned solutions are similar to GLRT in the Gaussian case, and signficantly better when the noise is non-Gaussian. In both cases, GLRT is theoretically CFAR, NET is not and CFAR-NET is approximately CFAR. }

% \label{fig:roc_unkown}%
% \end{figure*}

\begin{figure*}[h]
\centering
%\captionsetup{justification=centering}
     \includegraphics[width=0.285\textwidth]{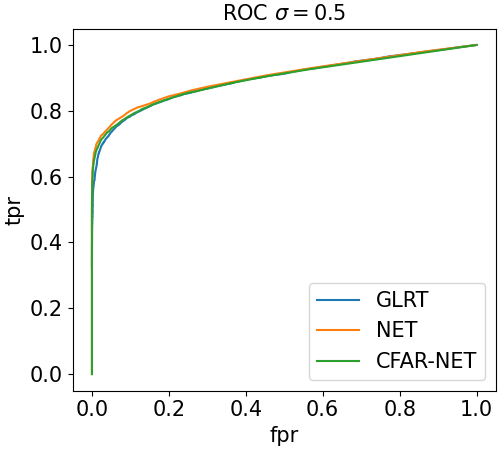}\includegraphics[width=0.285\textwidth]{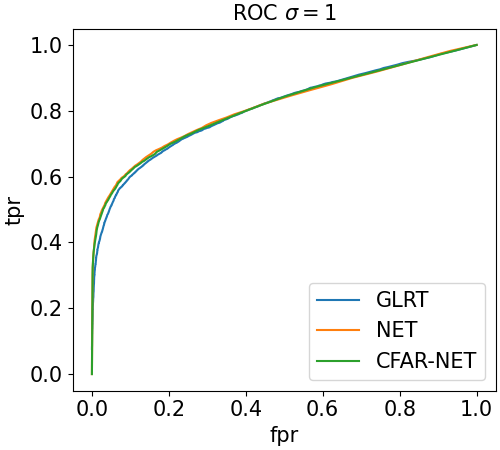}\includegraphics[width=0.285\textwidth]{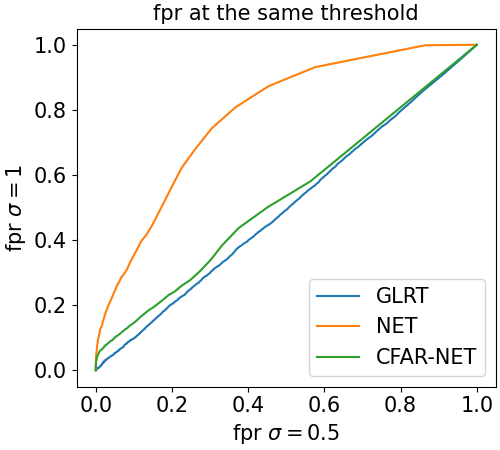} \\
     \includegraphics[width=0.285\textwidth]{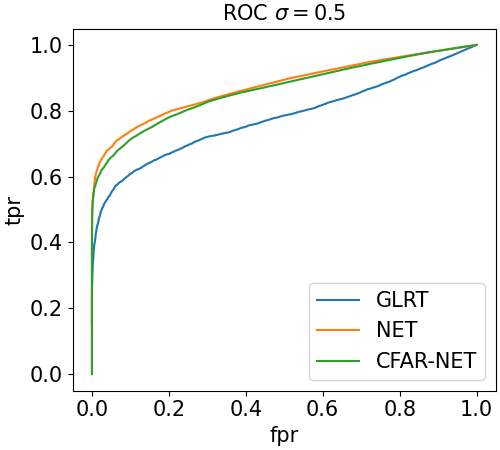}\includegraphics[width=0.285\textwidth]{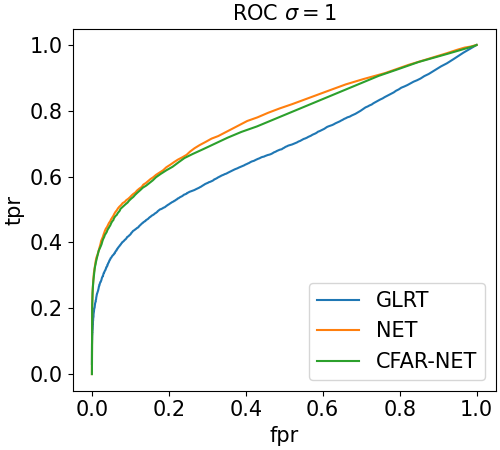}\includegraphics[width=0.285\textwidth]{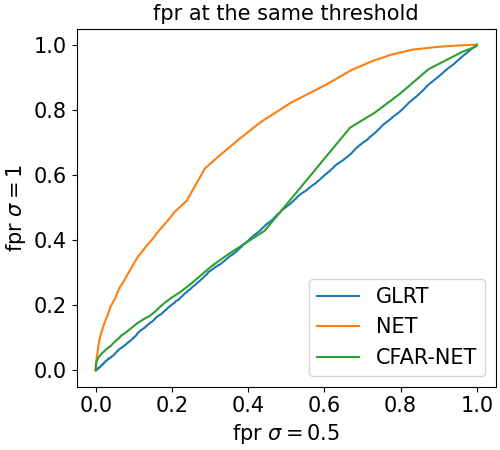}
\caption{Performance graphs in terms of FPR, TPR and thresholds for $\sigma\in\{0,1\}$. Top row is with Gaussian noise and indeed the Gaussian GLRT is best both in terms of ROC and CFAR. NET succeeds in achieving its ROC but is not CFAR. CFAR-NET is both accurate and CFAR. Bottom row is with non-Gaussian noise and the learned detectors beat the Gaussian GLRT in accuracy. CFAR-NET is near CFAR with a slight decrease in accuracy. }

\label{fig:roc_unkown}%
\end{figure*}

In this section, we introduce CFARnet, a framework for designing CFAR detectors using machine learning. CFARnet introduces two modifications to Algorithm 1. First, we augment the classification loss with a penalty function that ensures similar distributions for all values of $\z\in {\mathcal{H}}_0$. Second, in order to optimize this penalty, we generate multiple $\{\x_{ij}\}_{j=1}^M$ for each $\z_i$ and use them to empirically approximate the CFAR penalty.

% \begin{equation}\label{dpd}
%   \min_{\hat{{T}}\in {\mathcal{H}}} \frac 1N \sum_{i=1}^N L(\hat T(\x_i),\z_i)+ \alpha R_{\rm{CFAR}}(\hat{T})
% \end{equation}where $\alpha$ is a regularization hyper-parameter. The regularizer 

The main idea is adding a penalty to the objective function that promotes a CFAR. The penalty is defined as a sum of distance functions between the distributions of $\hat{T}$ under different values of $\z$:
\begin{align}\label{const2}
    R(\hat{T})=\sum_{\z, \tilde\z\in {\mathcal{Z}}_0} d\left(\hat{T}(\x); \hat{T}(\tilde\x)\right)
\end{align}
where $d(\cdot;\cdot)$ is any statistical distance as detailed in Section \ref{sec_dist}. The distributions of $\hat T(\x)$ and $\hat T(\tilde \x)$ are implicitly defined through
\begin{align}
    &\x\sim p(\x;\z)\nonumber\\
    &\tilde\x\sim p(\x;\tilde\z).
\end{align}
Clearly, any CFAR test must satisfy $R(\hat{T})=0$. 

To minimize a loss with a CFAR penalty, we need to represent each distribution using a small dataset. For each $\z_i$, we synthetically generate multiple observations $\x_{ij}$ for $j=1,\cdots,M$. Similarly, for each $\tilde\z_i$ we compute multiple $\tilde\x_{ij}$. We then plug these into the empirical distances as detailed in Sec. \ref{sec_dist}:
\begin{align}\label{const22}
    \hat R(\hat T)=\sum_{i,\tilde{i}} \hat d\left(\{\hat{T}(\x_{ij}\}_{j=1}^M); \{\hat{T}(\tilde\x_{\tilde{i}j})\}_{j=1}^M\right)
\end{align}
% Given the enhanced dataset defined above, the CFAR penalty can be simply approximated by comparing  empirical moments
% \begin{align}\label{Rk}
%     \hat R_k(\hat{T})= \sum_{i=1}^N\left|\sum_{j=1}^MT^k_{ij}-\sum_{j=1}^M\tilde{T}^k_{ij}\right|^2
% \end{align}
% More advanced metrics can be obtained using adversarial methods as 
% \begin{align}
%     \hat R_{\rm{adv}}(\hat{T})= \sum_{i=1}^N\max_{g(\cdot)}\left|\sum_{j=1}^Mg(T_{ij})-\sum_{j=1}^Mg(\tilde{T}_{ij})\right|^2
% \end{align}
% as common in generative adversarial networks (GANs) \cite{goodfellow2014generative,arjovsky2017wasserstein}. Either one of these approximations can be used as a penalty function in the fitting optimization.

Altogether, we recommend to use a hyper-parameter $\alpha>0$ that trades off the importance of the classification accuracy versus the CFAR penalty. The overall procedure for learning a CFAR detector is summarized in Algorithm 2.

\begin{algorithm}
\caption{Fitting CFAR detectors for $p(\x;\z)$}\label{alg:cfar}
\begin{itemize}
    \item Choose   $p^{\rm{fake}}(y)$.
    \item Choose  $p^{\rm{fake}}(\z;y)$.
    \item For each $i=1,\cdots,N$:
    
    \hspace{4mm}Generate $y_i$.
    
    \hspace{4mm}Generate $\z_i$ given $y_i$.
    
    \hspace{4mm}For $j=1,\cdots,M$:
    
    \hspace{8mm} Generate $\x_{ij}$ given $\z_i$.
    \item Solve 
    
    \hspace{4mm}$\min_{\hat{{T}}\in {\mathcal{T}}} \frac 1N \sum_{i=1}^N L(\hat T(\x_i),y_i)+\alpha \hat R(\hat T)$.
\end{itemize}
\end{algorithm}

% \begin{align}\label{const1}
%     {\mathcal{P}}_{\rm{asymp}}=\left\{\x\sim p(\x;\z)\;\rightarrow\;\hat{T}(\x)\sim T^{\rm{asymp}}_{\rm{GLRT}}(\x)\qquad \forall \quad \z\right\}
% \end{align}
%s defined in (\ref{asympP}). 

% For example, suppose we desire a CFAR detector where $\hat{T}$ behaves similarly under $\z_i$ and $\z_{i'}$, then we can use the following regularizer that penalizes unequal moments of order $k$:
% \begin{align}
%     R_k(\hat{T})=\sum_{i\neq i'} \left|\sum_j\hat{T}^k(\x_{ij})-\sum_j\hat{T}^k(\x_{i'j})^k\right|^2
% \end{align}
% The controlled-detector is then defined as the solution to
% \begin{equation}\label{controlled loss}
%     \min_{\hat{T}\in\mathcal{H}}\sum_{i}d\left(\left\{ \hat{T}\left(\x_{ij}\right)\right\}_{j=1}^{M};p^{\rm asymp}(2\log \left(T_{\rm GLRT}(\x)\right);\z_i)\right)
% \end{equation}
% where 
% \begin{align}
%     d(\{t_j\}_{j=1}^M;p_t(t))
% \end{align}
% is a distance function between the empirical distribution of $\{t_j\}_{j=1}^M$ and the desired distribution $p_t(t)$. Examples for scalable distances suitable for minimization include Generative Adversarial Networks \cite{goodfellow2014generative}, Wasserstein \cite{arjovsky2017wasserstein}, or moment matching distance (MMD) \cite{mom}.

\section{Numerical experiments}

In this section, we demonstrate the advantages of CFARnet via numerical experiments. We consider a basic yet realistic target detection scenario in which both the target amplitude and the noise scaling are unknown:
\begin{align}
    \x=A\1+\sigma\n
\end{align}
where $\1$ is a target vector of ones, $\n$ is a random vector with i.i.d. noise variables, and $\z=[A,\sigma]$ are deterministic unknown parameters 
\begin{align}\label{params}
    -1\leq A\leq 1,\qquad 0.5\leq \sigma\leq 1.
\end{align}
The goal is to decide between
\begin{align}
    &y=0:\quad A=0\nonumber\\
    &y=1:\quad A\neq 0.
\end{align}
We compare three detectors:
\begin{itemize}
    \item GLRT: assuming Gaussian noise, the classical GLRT has a simple closed form solution $T_{\rm{GLRT}}=(\x^T\1)^2/(\x^T\x)$ and is known to be CFAR.
    % \begin{align}
    %     GLRT(\x)=\frac{(\x^T\1)^2}{\x^T\x}
    % \end{align}
    \item NET: a learned neural network as in Algorithm 1. We choose a uniform fake prior for the unknown parameters in (\ref{params}).  The architecture is based on four  non-linear features: the sample mean of $\x$, its sample variance and robust versions of the two based on the median. These features are passed through a fully connected neural network, and are optimized to minimize a cross entropy loss using PyTorch. 
    \item CFAR-NET: a learned neural network as in Algorithm 2. Architecture and implementation are all identical to NET. Loss is cross entropy with an MMD CFAR penalty with parameter $\alpha=1$. 
\end{itemize}

The first experiment considers Gaussian noise. In the first row of Fig. 1, we plot the two ROCs for different values of $\sigma$. To examine the CFAR property, we also plot the FPRs under different parameters. As expected, it is easy to see that the Gaussian GLRT performs well and is CFAR. NET provides similar accuracy as illustrated in its ROC but is non-CFAR and results in significantly different FPR when we change $\sigma$. On the other hand, CFAR-NET is both accurate and near CFAR.

The second experiment is more challenging and considers non-Gaussian noise. The setting is identical as before except for the noise distribution
\begin{align}
    p(n_k)= (1-\epsilon)N(0,1) + \epsilon N(0,100) 
\end{align}
where $\epsilon=0.1$. There is no simple GLRT for this setting. The results are provided in the second row of Fig. 1. In this case, the Gaussian GLRT is no longer optimal and the two learned detectors provide a significantly better ROC. In terms of CFAR, GLRT is still invariant to the nuisance parameter, but the FPR of NET is dependent on its value. As promised, CFAR-NET is both accurate and CFAR. 

\section*{Acknowledgment}
The authors would like to thank Yoav Wald and Yiftach Beer for fruitful discussions and helpful insights in the initial stages of this project.

\bibliographystyle{ieeetr}
\bibliography{main.bib}

\end{document}